\documentclass{article}

\usepackage[preprint]{styles/corl_2026}
\usepackage{titletoc}
\usepackage{amsmath}
\usepackage{amssymb}
\usepackage{amsthm}
\usepackage{bm}
\usepackage{microtype}

\usepackage{graphicx}
\usepackage{booktabs}
\usepackage{multirow}
\usepackage{makecell}
\usepackage{subcaption}

\usepackage{algorithm}
\usepackage{algpseudocode}
\usepackage{algorithmicx}

\usepackage{enumitem}
\usepackage{xcolor}
\usepackage{url}
\usepackage{hyperref}
\hypersetup{
    colorlinks=true,
    linkcolor=blue!60!black,
    citecolor=green!50!black,
    urlcolor=blue!60!black,
}
\usepackage{cleveref}
\usepackage{wrapfig}
\usepackage[table]{xcolor}
\definecolor{rowblue}{RGB}{235, 242, 255}
\definecolor{rowgray}{RGB}{245, 245, 245}
\definecolor{rowcolor}{RGB}{235, 235, 255}

\newcommand{\gdown}[1]{\textcolor{green!60!black}{\scriptsize$\downarrow$#1}}

\theoremstyle{definition}

\theoremstyle{remark}

\title{\vspace{-0.3in} Finetuning Vision‑Language‑Action Models Requires Fewer Layers Than You Think}
\author{
  \parbox{\textwidth}{\centering
  \vspace{0.2in}
    Gia-Binh Nguyen$^{1,2}$ \quad
    Trong-Bao Ho$^{2}$ \quad
    Thien-Loc Ha$^{2}$ \quad
    Khoa Vo$^{3}$ \quad
    Philip Lund M{\o}ller$^{4}$ \\[0.3em]
    Quang T. Nguyen$^{2}$ \quad
    Long Dinh$^{1,2}$ \quad
    Tung M. Luu$^{5}$ \quad
    Tuan Dam$^{6}$ \quad
    Vu Duong$^{1}$ \quad
    Trung Le$^{7}$ \\[0.3em]
    Nghi D. Q. Bui$^{1}$\quad 
    Minh Vu$^{1,2}$ \quad
    Tran N. Le$^{4}$ \quad
    An T. Le$^{1,2,13}$ \quad
    Ngan Le$^{3}$ \quad 
    Daniel Sonntag$^{8,9}$ \\[0.3em]
    James Zou$^{12}$ \quad
    Jan Peters$^{9,13}$ \quad
    Duy M. H. Nguyen$^{\dagger,9,10,11}$ \quad
    Ngo Anh Vien$^{\dagger,1,2}$ \\[1.2em]
    \footnotesize\normalfont
    $^{1}$Center for AI Research, VinUniversity \quad
    $^{2}$VinRobotics \quad
    $^{3}$University of Arkansas \\[0.3em]
    $^{4}$Technical University of Denmark \quad
    $^{5}$KAIST \quad
    $^{6}$ Hanoi University of Science and Technology \\[0.3em]
    $^{7}$Monash University \quad
    $^{8}$Oldenburg University \quad
    $^{9}$DFKI \quad
    $^{10}$University of Stuttgart \\[0.3em]
    $^{11}$IMPRS-IS \quad 
    $^{12}$Stanford University \quad
    $^{13}$Technische Universität Darmstadt \\[0.4em]
    \footnotesize\normalfont $^{\dagger}$Project Leads.
  }
}
\begin{document}
\maketitle
\vspace{-0.2in}
\begin{abstract}
%%%%%%%% Duy's version %%%%%%%%%
Vision-Language-Action (VLA) models pre-trained on massive video-robot datasets have revolutionized robotic manipulation, yet their multi-billion parameter architectures impose prohibitive computational burdens during downstream fine-tuning and real-time inference. In this work, we reveal a highly non-trivial architectural characteristic of these continuous control foundation policies (e.g., $\pi_0$, GR00T-N1.5): despite being trained on diverse physical trajectories, they exhibit severe layer-wise representational redundancy. To exploit this, we introduce a structural compression pipeline that is entirely training-free, bypassing the need of existing methods to load full-scale models to learn optimized token reductions or dynamic layer selectors. Instead, using only a single forward pass via Centered Kernel Alignment to identify redundant layer features, we remove twin layers to permanently compress the model depth by up to 50\% across both the VLM backbone and the continuous control policy head. Downstream fine-tuning of this streamlined architecture yields a dual acceleration benefit: a 40–50\% reduction in training time and up to 30\% faster real-time inference, while matching or exceeding full-scale base model performance. We comprehensively validate our method across \textbf{three simulation benchmarks} (LIBERO, RoboCasa, SimplerEnv) and \textbf{10 diverse real-world manipulation tasks} across \textbf{4 unique robotic embodiments}. These results prove that advanced VLAs require significantly fewer layers than previously assumed, offering a highly compute-efficient paradigm for scalable robot learning. Project page: \href{https://clpvla.github.io/}{\textcolor{blue}{https://clpvla.github.io/}}

%%%%%%%%% Binh's version %%%%%%%%%
% Vision-Language-Action (VLA) models have achieved impressive results in robotic manipulation by leveraging large-scale transformer architectures, yet their heavy computational demands during finetuning and strict inference latency requirements pose critical challenges for real-world deployment. By analyzing the representation across transformer layers, we find that a large proportion of layers encode highly redundant representations and can be safely pruned before finetuning. Building on this finding, we propose a simple but effective pipeline -- \textbf{CKA-guided} pruning then finetuning that removes 30--60\% of transformer layers while preserving both task performance and internal representational behavior. Extensive experiments across multiple backbones ($\pi_0$, GR00TN1.5, SmolVLA) and simulation benchmark (LIBERO, RoboCasa, SimplerEnv) and 10 real-world manipulation tasks across 4 robot embodiments covering deformable and rigid objects, diverse skills, and different data scales from 100 to 2,800 demonstrations consistently demonstrate around 30\% reduction in trainable parameters, 30--50\% faster training, and nearly 30\% faster inference, while maintaining competitive performance with the full base model and even surpassing it on several tasks with limited data. Our results suggest that finetuning VLAs requires far fewer layers than commonly assumed, offering a practical and accessible path toward efficient robot learning at scale.

\end{abstract}


% ============================================================
\section{Introduction}
\label{sec:01_introduction}
Recent years have witnessed a major paradigm shift in robotic manipulation driven by the emergence of Vision-Language-Action (VLA) models. By framing physical control as a multimodal translation problem, early architectures like RT-2~\cite{zitkovich2023rt}, OpenVLA~\cite{kim2024openvla}, and CogACT~\cite{li2024cogact} demonstrated remarkable generalization capabilities across language-conditioned tasks. More recently, state-of-the-art (SOTA) architectures have transitioned toward continuous action generation using diffusion or flow-matching objectives, such as $\pi_0$~\cite{black2024pi_0} and the GR00T family~\cite{bjorck2025gr00t}, yielding significantly more robust trajectory generation, smoother control paths, and tighter physics-priors. However, these architectural \textit{breakthroughs are tightly coupled with massive scale}; modern generalist VLA policies regularly span \textit{billions of parameters}. This extreme scale introduces a severe computational bottleneck, manifesting as staggering hardware and cluster time costs during downstream training, alongside high operational latency and heavy memory-bandwidth overhead that strictly constrain real-time edge inference.

To alleviate these computational costs, the embodied AI community has explored several distinct optimization strategies.
One line of work accelerates action generation or visual processing through parallel decoding, decoupling speculative execution, token pruning, and temporal caching (training-free), including OpenVLA-OFT~\cite{kim2025fine}, Knowledge Insulating \cite{driess2026knowledge}, EfficientVLA~\cite{yang2026efficientvla}, SpecPrune-VLA~\cite{wang2025specprune}. 
Another line designs lighter VLA architectures from scratch, such as RoboMamba~\cite{liu2024robomamba}, Flower-VLA~\cite{reuss2025flower}, SmolVLA~\cite{shukor2025smolvla}, and NORA~\cite{hung2025nora}. These models improve accessibility, but may not fully retain the broad capabilities inherited from large-scale pretrained VLA backbones. A third direction modifies pretrained models through training-adaptive computation, such as DeeR-VLA~\cite{yue2024deer}and Mole-VLA~\cite{zhang2026mole}, which learn to exit early or dynamically route inputs through a subset of layers. While promising, these approaches often require auxiliary routing modules, additional training objectives, or task-dependent runtime decisions.

Despite promising efficiency gains, existing VLA optimization frameworks face three critical bottlenecks at the intersection of model compression and physical deployment. First, \textit{the architectural and evaluation scope remains narrow}: current techniques focus on older, text-token autoregressive baselines (e.g., OpenVLA) rather than modern continuous-control models ($\pi_0$, GR00T-N1.5), with real-world validation often restricted to 1–2 simple tasks~\cite{yang2026efficientvla,xu2026vla}. Second,\textit{ training-free methods fail to accelerate the expensive downstream fine-tuning phase}, which remains a major bottleneck in robot learning; for instance, a typical training on Libero often requires up to 20 hours on 4$\times$ A100 GPUs to reach convergence~\cite{nguyen2026foca,koo2025hamlet}. Third, while \textit{training-adaptive methods} enable faster training and inference, they \textit{introduce immense architectural complexity} via auxiliary routing subcomponents or distillation pipelines~\cite{zhang2026mole,yue2024deer}, fundamentally altering the core model structure and creating friction with downstream learning algorithms. Together, these challenges raise a key question: \textit{``Can redundant layers be removed before fine-tuning without sacrificing policy performance?"}

% \Ngan{I prefer to use italic instead of bold font} \Ngan{I suggest to conclude the current challenges by a scientific question which we will solve in the later sections: can we statically remove redundant backbone depth before fine-tuning, thereby reducing both adaptation and inference cost while preserving policy performance?}

We solve these challenges via a structurally clean and efficient framework that \textit{eliminates the need to maintain full backbone depth during fine-tuning}, specifically targeting modern continuous control foundation models such as $\pi_0$ and GR00T-N1.5. Our key observation is that consecutive transformer blocks in modern VLA backbones often produce highly correlated token representations, suggesting substantial depth-wise redundancy.
% The method is driven by the finding that consecutive blocks in modern continuous-control foundations generate highly correlated token representations that do not alter latent trajectory features. 
% \Ngan{This sentence is rewritten: Our key observation is that consecutive transformer blocks in modern VLA backbones often produce highly correlated token representations, suggesting substantial depth-wise redundancy}.
Specifically, we propose \textsc{CLP} (\textbf{C}KA-guided \textbf{L}ayer \textbf{P}runing), which performs a single forward pass and uses Centered Kernel Alignment (CKA)~\cite{kornblith2019similarity,cortes2012algorithms,nguyen2021do} to identify groups of representationally redundant layers. These layers are then permanently removed before fine-tuning. Unlike token pruning, caching, speculative decoding, or dynamic routing, \textsc{CLP} produces a statically smaller model that reduces active parameters, memory usage, training cost, and inference latency without adding new modules or auxiliary objectives.

% Specifically, by executing a single forward pass on a sample batch of data before training, we utilize Centered Kernel Alignment (CKA)~\cite{kornblith2019similarity,cortes2012algorithms,nguyen2021do} to geometrically detect groups of layer-wise redundancy and then permanently prune these redundant layer blocks before fine-tuning, namely \textsc{CLP} (\textbf{C}KA-guided \textbf{L}ayer \textbf{P}runing). This completely bypasses the intense architectural complexity or learning dynamic routing subcomponents as prior methods, and permanently scales down the active parameter space, yielding to slashing the multi-hour, multi-GPU training bottleneck alongside inference latency. \Ngan{The second last sentence is important but not clear. The last sentence is not suitable at introduction level. I suggest to emphasize on CLP and compare with the existing methods for a better motivation: Specifically, we propose \textsc{CLP}, which performs a single forward pass and uses Centered Kernel Alignment (CKA)~\cite{kornblith2019similarity,cortes2012algorithms,nguyen2021do} to identify groups of representationally redundant layers. These layers are then permanently removed before fine-tuning. Unlike token pruning, caching, speculative decoding, or dynamic routing, \textsc{CLP} produces a statically smaller model that reduces active parameters, memory usage, training cost, and inference latency without adding new modules or auxiliary objectives.}

We comprehensively validate our framework across three popular backbones ($\pi_0$, GR00T-N1.5, SmolVLA) over \textit{3 simulation benchmarks} (LIBERO~\cite{liu2023libero}, RoboCasa~\cite{nasiriany2024robocasa}, SimplerEnv~\cite{li2024simplerenv}) and \textit{10 demanding real-world manipulation tasks} spanning \textit{4 distinct robotic embodiments} (Aloha Single and bimanual arms~\cite{zhao2023aloha}, UR10 and UR5~\cite{hawkins2013analytic,universalrobotsmanual}). In simulation, our method eliminates roughly 30\% of trainable parameters and training time while maintaining competitive success rates. In real-world physical deployments, our streamlined architecture matches or even surpasses full-scale baselines. Crucially, this structural compression acts as an effective regularizer in \textit{data-scarce regimes}: restricting training to only 10\% of LIBERO data lifts success rates from 77.7\% to 84.6\%, and on real-world tasks with limited datasets of just 100 demonstrations, it delivers a \textit{15\% to 20\% performance boost} over full-scale models.

Our main contributions are summarized as follows:
\begin{itemize}[leftmargin=*, itemsep=3pt, topsep=0pt, parsep=0pt, partopsep=2pt]
    \item \textbf{Accessible Adaptation of SOTA VLAs:} We demonstrate that SOTA continuous-control foundations (such as $\pi_0$ and GR00T-N1.5) can be fine-tuned with substantially reduced depth, lowering memory, training, and inference cost without introducing auxiliary routing or early-exit modules. 
    \item \textbf{Pre-Finetuning Backbone Compression}. To this end, we propose \textsc{CLP}, a simple calibration-based framework that identifies representationally redundant transformer blocks using CKA and removes them before downstream adaptation.
    % \item To this end, we propose \textsc{CLP}, a simple yet effective CKA-guided pruning framework that identifies and removes representationally redundant transformer layers prior to downstream adaptation.
    % can be adapted using significantly fewer layers without introducing complex sub-modules into the architecture. This provides a simple, accessible pathway that efficiently lowers computational costs for the research community, while raising fundamental questions about the systemic over-parameterization of these models during their initial pre-training phase. \Ngan{shorten the second part: ....  and GR00T-N1.5) can be fine-tuned with substantially reduced depth, lowering memory, training, and inference cost without introducing auxiliary routing or early-exit modules.}   
    % \item \Ngan{I prefer to include CLP as one of the main contributions as it is the heart of your method: We introduce \textsc{CLP}, a simple calibration-based framework that identifies representationally redundant transformer blocks using CKA and removes them before downstream adaptation.}
    % \item \Ngan{I suggest to compare the last two concept into one contribution as broad multi-embodiment validation: we evaluate our framework cross three simulation benchmarks, ten real-world manipulation tasks, and four robotic embodiments, demonstrating that static depth reduction preserves performance and improves sample efficiency in low-data regimes.}
    \item \textbf{Diversified Multi-Embodiment Validation}: We conduct extensive evaluations across multiple simulation benchmarks and 10 real-world manipulation tasks spanning four robotic embodiments, demonstrating that our non-invasive depth reduction is robust, platform-agnostic, and highly generalizable. Beyond achieving competitive performance against prior acceleration methods, CLP consistently improves sample efficiency under limited-data regimes, acting as an effective structural regularizer.
\end{itemize}
% \Ngan{include SmolVLA and NORA for the existing work} \Ngan{emphasize on the question "can we statically remove redundant backbone depth before fine-tuning to reduce both adaptation cost and deployment cost while preserving policy behavior?", which was limited in the existing work}
% \textcolor{orange}{duy:updated!}

% ============================================================
\section{Related Work}
\label{sec:02_related_work}
% --------- Binh's version
% \paragraph{Vision-Language-Action Models.}
% Vision-Language-Action (VLA) models unify visual perception, text comprehension, and robotic control for general manipulation. Current frameworks generally follow two paradigms: autoregressive token generation, where models like RT-1~\cite{brohan2022rt}, RT-2~\cite{zitkovich2023rt}, OpenVLA~\cite{kim2024openvla}, and SpatialVLA~\cite{qu2025spatialvla} output discrete action tokens, and diffusion-based continuous sequence generation—such as $\pi_0$~\cite{black2024pi_0}, Octo~\cite{team2024octo}, and the GR00T family~\cite{bjorck2025GR00T}—which yields smoother physical control. Both paradigms rely heavily on deep transformer backbones whose massive parameter counts introduce severe computational overhead during downstream training and real-time inference.

% Motivated by these constraints, we find that many consecutive layers in VLA backbones yield highly correlated hidden-state representations. Based on this observation, our \textbf{CKA-Guided} pruning framework identifies and permanently deletes redundant blocks prior to adaptation. This structural reduction significantly cuts training costs and inference latency while maintaining competitive performance and preserving original internal behaviors without relying on knowledge distillation or auxiliary losses.

\vspace{-0.1in}
% Updated version
\textbf{VLA Models and Efficient Architectures.} 
% Current frameworks in VLA generally follow two paradigms: autoregressive token generation, where models like RT-1~\cite{brohan2022rt}, RT-2~\cite{{zitkovich2023rt}}, OpenVLA~\cite{kim2024openvla}, and SpatialVLA~\cite{qu2025spatialvla} output discrete action tokens, and diffusion- or flow-matching-based continuous sequence generation such as Octo~\cite{team2024octo}, CogACT~\cite{li2024cogact} $\pi_0$~\cite{black2024pi_0}, and the GR00T family~\cite{bjorck2025GR00T}, which yields smoother physical control and be one of strongest models. While both paradigms demonstrate remarkable generalization capabilities, their reliance on deep transformer backbones introduces severe computational overhead. This motivates a growing line of research that focuses on building inherently smaller yet capable VLA models from scratch. 
Current VLA frameworks primarily split into two paradigms: autoregressive architectures that output discrete action tokens (e.g., RT-1/2, OpenVLA, SpatialVLA) and diffusion- or flow-matching-based models that generate continuous control sequences (e.g., Octo, CogACT, $\pi_0$, GR00T-N1.5). While both exhibit exceptional generalization, their deep transformer backbones incur severe computational overhead. This bottleneck has motivated a growing line of research focused on training inherently smaller, high-capability VLA models from scratch.
For instance, RoboMamba~\cite{liu2024robomamba} integrates linear-complexity State Space Models (SSMs)~\cite{gu2023mamba} to achieve a multi-fold speedup in inference over attention-heavy baselines. Concurrently, lightweight frameworks like FLOWER-VLA~\cite{reuss2025flower}, SmolVLA~\cite{shukor2025smolvla} and NORA~\cite{hung2025nora} leverage compact multi-modal backbones (such as Florence-2~\cite{xiao2024florence2} or Qwen-2.5-VL~\cite{qwen2025vl}) and optimized action-token designs to maximize task performance while maintaining a restricted parameter footprint.

While these approaches improve accessibility, they typically require training or designing a new compact model from scratch and may not fully inherit the broad capabilities of large pretrained VLA foundations. In contrast, our method is complementary: we directly compress existing pretrained VLAs by removing redundant transformer layers before downstream fine-tuning.

\textbf{Acceleration Techniques for VLA Model.}
Recent VLA efficiency methods can be further categorized into 
% \Ngan{Recent VLA efficiency methods can be further categorized into}
:\textit{(i) Training-free method} and \textit{(ii) training-adaptive ones}. The first one optimizes computational density by filtering token sequences using multi-modal and kinetic cues: VLA-Cache~\cite{xu2026vla} exploits temporal frame continuity to adaptively cache and reuse token KV states~\cite{ge2024model,zhou2024dynamickv}; Efficient-VLA~\cite{yang2026efficientvla} distills a compact, spatially diverse token subset prioritized by task relevance; SpecPrune-VLA~\cite{wang2025specprune} executes a speculative dual-level pruning that eliminates background tokens using historical attention and depth-dependent importance; and ADP~\cite{pei2025action} dynamically gates token dropping by coupling text semantics directly with real-time end-effector kinetics. Collectively, these frameworks shift the computational burden by adaptively matching token processing to the immediate demands of the manipulation environment.

% In contrast, \textit{training-adaptive frameworks} \textit{structurally modify or expand the network} to enforce computational sparsity, requiring targeted optimization phases to balance architectural efficiency with raw policy performance. For instance, MoE-VLA~\cite{zhang2026mole} integrates sparse Mixture-of-Experts~\cite{shazeer2017outrageously,fedus2022switch,riquelme2021scaling} routing layers to conditionally activate specialized token- or action-level sub-networks tailored to specific linguistic task instructions. Taking a depth-wise approach, DeeR-VLA~\cite{yue2024deer} embeds multiple intermediate prediction heads to enable dynamic early-exiting, terminating forward propagation entirely when a less complex environmental state allows for a shallow action readout. Currently, $\text{AC}^2$-VLA~\cite{yu2026ac} introduces an action-prior router trained via action-guided self-distillation to jointly select token dropping and layer skipping conditioned on historical action context. While highly potent, these paradigms mandate specialized architectural modifications and training overhead to properly learn and stabilize their dynamic routing mechanisms.

Conversely, training-adaptive frameworks structurally modify or expand the network to enforce computational sparsity, balancing efficiency and policy performance through targeted optimization. For instance, MoLe-VLA~\cite{zhang2026mole} integrates sparse routing layers to activate task-specific sub-networks. Moving depth-wise, DeeR-VLA~\cite{yue2024deer} embeds intermediate prediction heads to enable dynamic early-exiting for simpler environmental states, while $\text{AC}^2$-VLA~\cite{yu2026ac} employs an action-prior router to jointly govern token dropping and layer skipping. Though potent, these methods require substantial architectural modifications and training overhead to stabilize their dynamic routing mechanisms.

% The \textsc{CLP} offers acceleration for both downstream training and real-time inference - a dual benefit that inference-only, training-free methods cannot provide. Furthermore, by avoiding multi-stage or sub-task distillation phases, our method provides a straightforward way to integrate advanced adaptation into VLA, such as future knowledge prediction and chain-of-thought, and is highly effective in low-data regimes, where training additional adaptive components is notoriously sensitive and prone to overfitting.

The \textsc{CLP} accelerates both downstream training and real-time inference, unlike inference-only training-free methods. By avoiding multi-stage distillation or auxiliary adaptive modules, such as the training-adaptive one, it also enables simple integration of advanced VLA adaptation, such as future knowledge prediction~\cite{nguyen2026foca,team2025gemini} and chain-of-thought reasoning~\cite{huang2026thinkact,zhang2026dreamvla}, while remaining highly effective in low-data regimes where additional trainable components often overfit (Figure\ref{fig:result_abcde}-c). 
% \Ngan{highlight this with \emph{}}

% ============================================================
\vspace{-0.1in}
\section{Preliminaries}
\begin{figure}[!htb]
    \centering
    \includegraphics[width=1.0\linewidth]{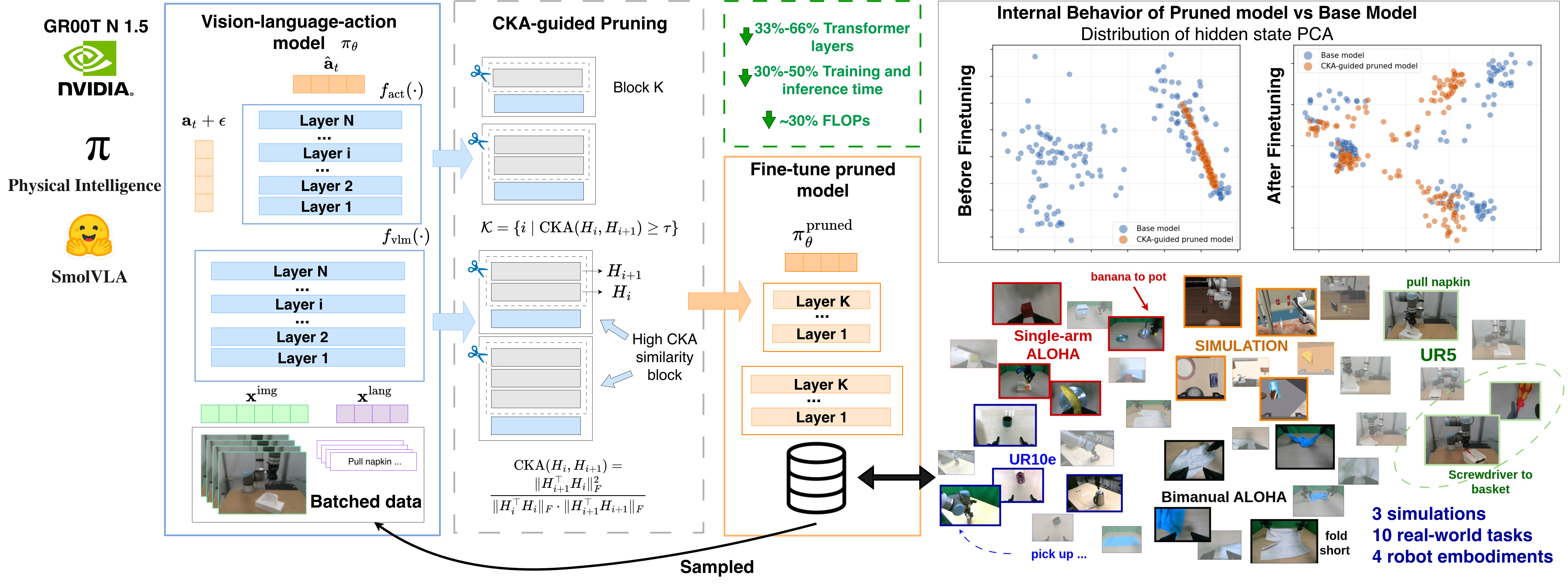}
    \caption{\small{\textbf{Overview of the proposed CLP framework.} CLP prunes representationally redundant transformer layers via CKA, reducing network depth by up to 66\% and training/inference cost by up to 50\%. Fine-tuning restores the latent geometry of the compressed model, enabling competitive performance across three simulation benchmarks, 10 real-world tasks, and four robotic embodiments.}}
    \vspace{-0.15in}
 \label{fig:method}
\end{figure}
\label{sec:03_preliminaries}
\vspace{-0.1in}
\textbf{Vision-Language-Action Model.}
% \Ngan{This section should be shorten, keep the key concepts, regarding notations used in the following sections. Otherwise, move to supplementary}
Let $\mathbf{x}^{\text{lang}}$ denote a text instruction and $\mathbf{x}^{\text{img}} \in \mathbb{R}^{H \times W \times 3}$ denote the RGB observation. A continuous-action VLA policy predicts an action chunk $\mathbf{a}\in\mathbb{R}^{T_a\times d_a}$ conditioned on these multimodal inputs. State-of-the-art continuous-control foundations (e.g., $\pi_0$~\cite{black2024pi_0}, GR00T~\cite{bjorck2025gr00t}, SmolVLA~\cite{shukor2025smolvla}) generally share a decoupled architecture: a Vision-Language Model (VLM) backbone that extracts environmental context, followed by a flow- or diffusion-based action-generation head.

% \begin{figure}[!htb]
%     \centering
%     \includegraphics[width=\linewidth]{images/method3.png}
%     \caption{CKA-Guided pruning}
%  \label{fig:method}
% \end{figure}

\noindent\underline{\textit{VLM Backbone.}} Following modern efficiency conventions~\cite{pei2025action}, the VLM backbone is modeled as a stack of $N_v$ transformer layers. Let $H^{\text{vlm}}_\ell$ denote the hidden token representation after the $\ell$-th VLM layer, and let $F^{\text{vlm}}_\ell$ denote the corresponding transformer block. The input hidden state $H^{\text{vlm}}_0$ is obtained by embedding the language and visual observations, and the final context representation is $Z = H^{\text{vlm}}_{N_v}$:
\begin{equation}
    H^{\text{vlm}}_0 = \operatorname{Embed}\left(\mathbf{x}^{\text{lang}}, \mathbf{x}^{\text{img}}\right), \quad H^{\text{vlm}}_\ell = F^{\text{vlm}}_\ell\left(H^{\text{vlm}}_{\ell-1}\right) \quad \forall \ell \in \{1,\ldots,N_v\} .
\end{equation}
\noindent\underline{\textit{Action-Generation Head.}} For flow-matching-based generation~\cite{lipman2023flow}, the action head parameterizes a velocity field to transport a Gaussian noise vector $\epsilon \sim \mathcal{N}(\mathbf{0},\mathbf{I})$ toward the target action $\mathbf{a}$ along a linear interpolation path $\mathbf{a}_t = (1-t)\epsilon + t\mathbf{a}$ for time step $t\in[0,1]$. Let $H^{\text{act}}_m$ denote the hidden action-token representation after the $m$-th action layer, and let $F^{\text{act}}_m$ denote the corresponding transformer block in an action head with $N_a$ layers. The forward pass is formalized as:
\begin{equation}
    H^{\text{act}}_0 = \operatorname{Embed}_{\text{act}}\left(\mathbf{a}_t,t\right), \quad H^{\text{act}}_m = F^{\text{act}}_m\left(H^{\text{act}}_{m-1};\, \Phi_m(Z)\right) \quad \forall m \in \{1,\ldots,N_a\} ,
\end{equation}
where $\Phi_m(Z)$ represents the cross-conditioning signal derived from the VLM context $Z$ and injected into the $m$-th action layer (e.g., via decoder cross-attention or token prefixing). 

The final layer output designates the predicted velocity field $\hat{\mathbf{u}}_t = f_{\text{act}}(Z, \mathbf{a}_t, t) = H^{\text{act}}_{N_a}$, which is optimized via the Flow Matching objective:
$  \mathcal{L}_{\text{FM}} = \mathbb{E}_{t,\mathbf{a},\epsilon} \left[ \left\| f_{\text{act}}\left(Z,\mathbf{a}_t,t\right) - \left(\mathbf{a}-\epsilon\right) \right\|_2^2 \right] .
$
Despite varying cross-conditioning strategies ($\Phi_m$) across baselines, this shared formulation frames both the VLM and action modules as deep transformer blocks, making their intermediate hidden states structurally compatible for layer-wise similarity analysis.

\paragraph{Centered Kernel Alignment.}
CKA~\cite{kornblith2019similarity} quantifies representation similarity between layers while remaining invariant to orthogonal transformations and isotropic scaling, making it ideal for identifying structural redundancies~\cite{gromov2024unreasonable, men2025shortgpt, chen2024streamlining}. Given hidden states $H_i, H_j \in \mathbb{R}^{n \times d}$ across $n$ tokens and $d$ dimensions, the alignment of their Gram matrices ($K = HH^\top$) via the centered Hilbert-Schmidt Independence Criterion ($\operatorname{HSIC}$) reduces for centered linear kernels to a ratio of Frobenius norms:
\setlength{\abovedisplayskip}{4pt}       % Space above equation
\setlength{\belowdisplayskip}{4pt}       % Space below equation
\setlength{\abovedisplayshortskip}{2pt}  % Space above if the preceding line is short
\setlength{\belowdisplayshortskip}{2pt}  % Space below if the preceding line is short
\begin{equation}
\operatorname{CKA}(H_i,H_j) = \frac{\operatorname{HSIC}(K_i,K_j)}{\sqrt{\operatorname{HSIC}(K_i,K_i)\cdot\operatorname{HSIC}(K_j,K_j)}} = \frac{\left|H_j^\top H_i\right|_F^2}{\left|H_i^\top H_i\right|_F \cdot \left|H_j^\top H_j\right|_F}.
\end{equation}
The score bounded within $[0,1]$ indicates representational similarity as it approaches $1$. In our framework, a high CKA score between adjacent VLA layers signals minimal feature transformation, rendering those blocks prime candidates for structured pruning.

% ============================================================
\section{CKA-Guided Layer Pruning}
\label{sec:04_redundant_layers}
\vspace{-0.1in}
\textbf{Representational Plateaus: Diagnosing Layer Redundancy in Deep VLAs.}
% To dismantle the extreme computational burden of deep VLAs, we first audit a fundamental premise: does massive scale translate to continuous representational progression? Leveraging the CKA framework formalized in Section~\ref{sec:03_preliminaries}, we inspect the hidden state trajectories across premier continuous-control models to isolate exactly where multi-modal context transforms into physical actions.
To better understand the internal mechanics driving deep VLA architectures, we initiate an empirical exploration into a central question: \textit{``how does information actually evolve across network depth?"}.
% \Ngan{use ``'' }
Leveraging the CKA metric established in Section~\ref{sec:03_preliminaries}, we trace the hidden state trajectories of representative continuous-control foundations as they map multimodal context to physical actions. This tracking reveals a striking structural phenomenon: rather than progressing uniformly or incrementally, the feature representations exhibit distinct zones of stagnation, uncovering widespread layer redundancy across both the backbone and the action head.

\begin{figure}[!htb]
    \centering
    \includegraphics[width=1.0\linewidth]{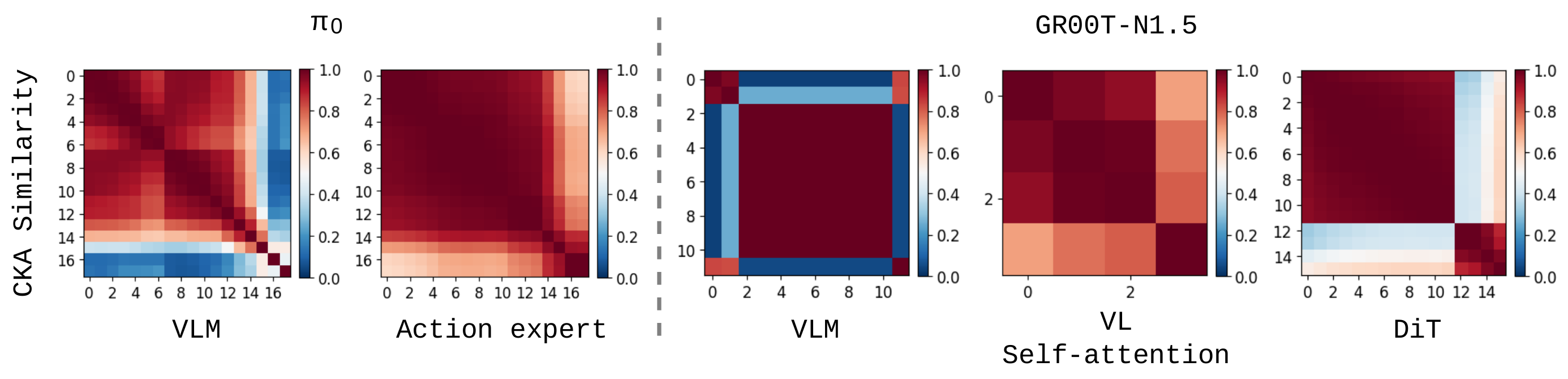}
    % \caption{CKA similarity among transformer layers in $\pi_0$ and GR00T-N1.5.}
    \vspace{-0.1in}
    \caption{\small{\textbf{CKA similarity profiles across $\pi_0$ and GR00T-N1.5 sub-modules.} The heatmaps illustrate pairwise representation alignment among transformer layers inside the VLM backbones, action heads, and DiT blocks. The extensive, contiguous plateaus of high similarity (dark red) across both model families signify minimal representational changes between successive layers, pinpointing candidate zones for structured pruning.}}
    % \vspace{-0.2in}
    % \vspace{0.4in}
 \label{fig:cka}
\end{figure}
\vspace{-0.2in}
% As illustrated in Figure~\ref{fig:cka}, this behavior manifests in both $\pi_0$ and GR00T-N1.5 as massive, contiguous plateaus of near-identity inter-layer similarity. This pervasive \textit{representational inertia} suggests that consecutive layers within these blocks perform highly redundant operations, while critical semantic shifts are compressed into a few sharp, localized transitions.

% This empirical landscape transforms a generic efficiency concern into an actionable structural hypothesis: these high-similarity blocks pinpoint latent structural slack, indicating that groups of adjacent layers can potentially be pruned without fracturing the underlying policy logic. While high CKA scores provide a robust geometric roadmap for identifying these candidate zones, they do not inherently guarantee downstream stability. We therefore utilize these profiles as a targeted diagnostic to isolate block candidates, verifying their practical dispensability through structured pruning and low-data fine-tuning.

As illustrated in Figure~\ref{fig:cka}, this behavior manifests in both $\pi_0$ and GR00T-N1.5 as large, contiguous blocks of high inter-layer similarity, indicating that consecutive layers perform highly redundant operations while major feature transformations are concentrated within a few distinct transitions. This pervasive redundancy motivates a concrete structural hypothesis: \textit{these high-similarity regions represent layers that can potentially be removed without degrading downstream policy performance}. However, because high representational similarity alone does not inherently guarantee stability, we only utilize these CKA profiles as an empirical diagnostic to isolate candidate blocks, verifying their pruning viability through structured layer removal and subsequent low-data fine-tuning.

\vspace{0.05in}
\textbf{CKA-Guided Pruning.}
Given a pre-trained VLA policy, we target the structured removal of transformer layers within a prunable module $\mathcal{M}$ (e.g., the VLM backbone or action head). Let $\mathcal{I}_{\mathcal{M}}=\{1,\ldots,L_{\mathcal{M}}\}$ index the ordered layers of $\mathcal{M}$. For a target pruning budget $k_{\mathcal{M}}$, we isolate a removal set $\mathcal{R}_{\mathcal{M}}\subset\mathcal{I}_{\mathcal{M}}$ ($|\mathcal{R}_{\mathcal{M}}|=k_{\mathcal{M}}$) to construct the compressed policy $\pi_{\theta}^{\mathrm{pruned}} = \operatorname{RemoveLayers}(\pi_\theta,\mathcal{R}_{\mathcal{M}})$. Because all transformer blocks within $\mathcal{M}$ share identical hidden dimensions, layer removal simply reconnects the remaining predecessor and successor blocks. This design enables direct fine-tuning under the native training objective without requiring auxiliary routing parameters, distillation losses, or architectural modifications.

We estimate representational redundancy using a compact calibration set $\mathcal{D}_{\mathrm{cal}}$ sampled from training episodes. By executing a forward pass over $\mathcal{D}_{\mathrm{cal}}$, we extract and concatenate the token representations across calibration examples to construct a unified layer activation matrix $\bar{H}^{\mathcal{M}}_\ell$ for each $\ell\in\mathcal{I}_{\mathcal{M}}$. We quantify sequential redundancy by computing CKA between consecutive layers:
\vspace{0.05in}
\begin{equation}
    s^{\mathcal{M}}_\ell = \operatorname{CKA}\left(\bar{H}^{\mathcal{M}}_{\ell-1},\bar{H}^{\mathcal{M}}_{\ell}\right), \quad \ell=2,\ldots,L_{\mathcal{M}} .
    \label{eq:cka_redundancy_score}
\end{equation}
An elevated similarity score $s^{\mathcal{M}}_\ell \approx 1.0$ indicates that layer $\ell$ introduces minimal representational change relative to layer $\ell-1$, marking it as a candidate for structured pruning. To prevent the disjointed removal of isolated layers due to local sampling noise, we aggregate adjacent redundant layers into contiguous blocks. Successive layers are clustered into the same block if they satisfy:
$
    s^{\mathcal{M}}_{\ell} \geq \tau, \quad \ell=2,\ldots,L_{\mathcal{M}},
    \label{eq:cka_block_threshold}
$
where $\tau$ is a designated similarity threshold. This constraint partitions the module into a set of contiguous high-similarity blocks $\mathcal{B}_{\mathcal{M}}=\{B_1,\ldots,B_Q\}$. For each block $B$, we retain its initial layer $r(B)$ as a functional anchor and pool the remaining layers into a candidate pruning set:
\setlength{\abovedisplayskip}{4pt}       % Space above equation
\setlength{\belowdisplayskip}{4pt}       % Space below equation
\setlength{\abovedisplayshortskip}{2pt}  % Space above if the preceding line is short
\setlength{\belowdisplayshortskip}{2pt}  % Space below if the preceding line is short
\begin{equation}
    \mathcal{P}_{\mathcal{M}} = \bigcup_{B\in\mathcal{B}_{\mathcal{M}}} \left(B\setminus\{r(B)\}\right).
\end{equation}
We calibrate $\tau$ to ensure the candidate pool satisfies $|\mathcal{P}_{\mathcal{M}}|\geq k_{\mathcal{M}}$. The final removal set $\mathcal{R}_{\mathcal{M}}$ is constructed by isolating the $k_{\mathcal{M}}$ most redundant layers within this pool:
\begin{equation}
    \mathcal{R}_{\mathcal{M}} = \operatorname{TopK}_{\ell\in\mathcal{P}_{\mathcal{M}}}\left(s^{\mathcal{M}}_\ell,k_{\mathcal{M}}\right) .
    \label{eq:topk_pruning}
\end{equation}
This targeted selection process is executed independently across each prunable module and formalized in Algorithm~\ref{alg:cka_pruning} (Appendix). The pruning routine is entirely static, permanently truncating network depth prior to downstream adaptation. Consequently, it shrinks both the fine-tuning training footprint and operational inference latency without introducing runtime computational overhead. 
% This empirical framework motivates two central questions: (i) what compression ratio can be sustained before policy performance degrades? and (ii) does the specific spatial localization of the pruned layers impact downstream policy stability?
\vspace{-0.8em}
% \begin{algorithm}[!hbt]
% \caption{CKA-Guided Layer Pruning}
% \label{alg:cka_pruning}
% \begin{algorithmic}[1]
% \Require Pre-trained policy $\pi_\theta$, target module $\mathcal{M}$, calibration set $\mathcal{D}_{\mathrm{cal}}$, budget $k_{\mathcal{M}}$, threshold $\tau$
% \State Extract calibrated hidden representations $\{\bar{H}^{\mathcal{M}}_1,\ldots,\bar{H}^{\mathcal{M}}_{L_{\mathcal{M}}}\}$ from $\mathcal{M}$ over $\mathcal{D}_{\mathrm{cal}}$
% \For{$\ell=2$ to $L_{\mathcal{M}}$}
%     \State $s^{\mathcal{M}}_\ell \leftarrow \operatorname{CKA}(\bar{H}^{\mathcal{M}}_{\ell-1},\bar{H}^{\mathcal{M}}_{\ell})$
% \EndFor
% \State Group consecutive layers into blocks $\mathcal{B}_{\mathcal{M}}$ where $s^{\mathcal{M}}_\ell\geq\tau$
% \State Identify candidate pool $\mathcal{P}_{\mathcal{M}}\leftarrow \bigcup_{B\in\mathcal{B}_{\mathcal{M}}}(B\setminus\{r(B)\})$, where $r(B)$ is the initial layer of block $B$
% \State Select removal set $\mathcal{R}_{\mathcal{M}}\leftarrow \operatorname{TopK}_{\ell\in\mathcal{P}_{\mathcal{M}}}(s^{\mathcal{M}}_\ell,k_{\mathcal{M}})$
% \State $\pi_\theta^{\mathrm{pruned}}\leftarrow \operatorname{RemoveLayers}(\pi_\theta,\mathcal{R}_{\mathcal{M}})$
% \State \Return $\pi_\theta^{\mathrm{pruned}}$
% \end{algorithmic}
% \end{algorithm}

% ============================================================
\section{Experimental Results and Analysis}
\label{sec:experiments}
\vspace{-0.1in}
We conduct a diverse set of experiments to examine the performance of pruned VLA models trained with our \texttt{CLP} models across simulation and real-world settings and robot embodiment configurations. Details are presented in the Appendix. We first investigate the fundamental architectural properties of layer redundancy in VLAs and dissect the internal mechanisms of our framework (\textbf{RQ1} \& \textbf{RQ2}). We then benchmark our method against standard baselines on large-scale manipulation suites (\textbf{RQ3}) and validate its physical feasibility on real-world robot platforms (\textbf{RQ4}):
\begin{itemize}[leftmargin=*, itemsep=3pt, topsep=0pt, parsep=0pt, partopsep=2pt]
    \item \textbf{RQ1 (Compression Trade-offs)}: To what extent can modern VLAs be structurally truncated before policy performance degrades?
    \item \textbf{RQ2 (Latent Behavior \& Ablation)}: Is CKA uniquely optimal for identifying functional redundancy, and how does downstream adaptation reshape the pruned latent space?
    \item \textbf{RQ3 (Baseline Comparison)}: How does our static pruning framework compare against state-of-the-art training-free and training-adaptive baselines across varied data regimes?
    \item \textbf{RQ4 (Real-World Deployment)}: Do the computed FLOP savings translate directly to wall-clock training speedups and high-frequency real-time inference on physical hardware?
\end{itemize}

\begin{figure*}[t]
    \centering
    \includegraphics[width=0.95\textwidth]{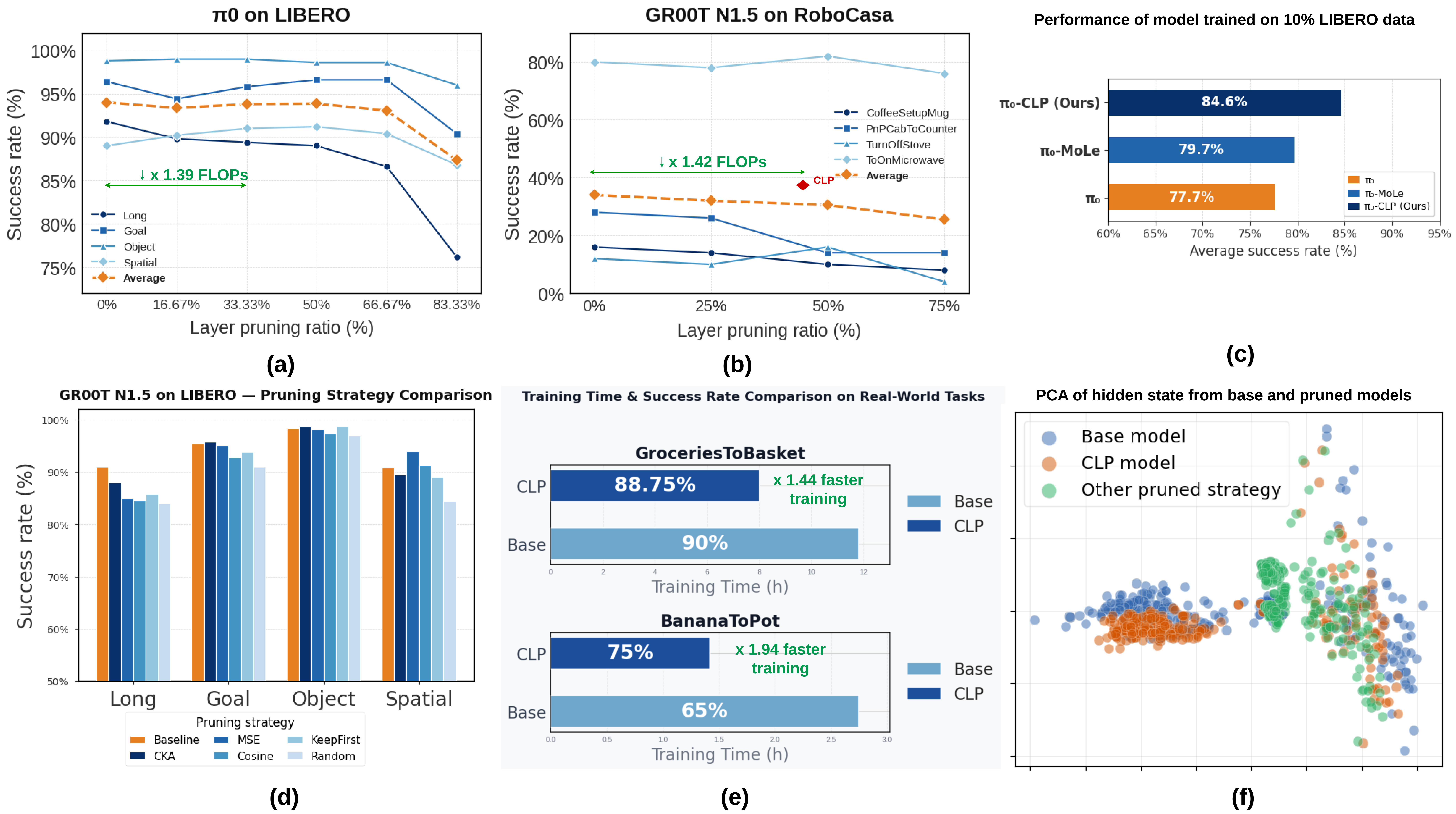}
    \caption{\small{
        \textbf{Analysis of CLP evaluation across benchmarks and real-world tasks.}
        \textbf{(a)} Success rate of $\pi_0$ on LIBERO with different layer pruning ratio;
        \textbf{(b)} Success rate of GR00T N1.5 on RoboCasa across pruning ratios;
        \textbf{(c)} Comparison with dynamic layer skipping method (MoLe-VLA~\cite{zhang2026mole}) vs ours;
        \textbf{(d)} Comparison of different pruning strategies on GR00T N1.5 across LIBERO benchmark;
        \textbf{(e)} Training time and success rate on real-world manipulation tasks;
        \textbf{(f)} PCA visualization of hidden states (state/future tokens and action tokens) for the Base model, CKA-guided pruned model with different pruning strategies.}
    }
    \vspace{-0.15in}
    \label{fig:result_abcde}
\end{figure*}
\noindent

\vspace{-0.2in}
\begin{table}[!htb]
\small
\renewcommand{\arraystretch}{0.9}
\setlength{\tabcolsep}{4pt}
\centering
\caption{\textbf{Efficiency comparison across models.} Model size, trainable parameters, training time (60000 steps), FLOPs, and inference speed on RTX 4070 reported on Libero benchmark.}
\vspace{0.05in}
\label{tab:efficiency}
\resizebox{\linewidth}{!}{
\begin{tabular}{c|c>{\columncolor{rowcolor}}c|c>{\columncolor{rowcolor}}c|c>{\columncolor{rowcolor}}c|c>{\columncolor{rowcolor}}c|c>{\columncolor{rowcolor}}c}
\toprule
\multirow{2}{*}{\textbf{Model}}
  & \multicolumn{2}{c|}{\textbf{Model Size}}
  & \multicolumn{2}{c|}{\textbf{Trainable Params}}
  & \multicolumn{2}{c|}{\textbf{Training Time (hours)}}
  & \multicolumn{2}{c|}{\textbf{GFLOPs}}
  & \multicolumn{2}{c}{\textbf{Inference Speed (ms)}} \\
  & Base & CLP & Base & CLP & Base & CLP & Base & CLP & Base & CLP \\
\midrule
$\pi_0$
  & 3.5B & 2.7B {\gdown{22.9\%}}
  & 3.1B & 2.3B {\gdown{25.8\%}}
  & 15.5 & 11.2 {\gdown{27.8\%}}
  & 3073 & 2196.5 {\gdown{28.5\%}}
  & 211 & 152 {\gdown{27.9\%}} \\
GR00TN1.5
  & 2.7B & 2B {\gdown{25.9\%}}
  & 1.07B & 0.75B {\gdown{30.1\%}}
  & 10.7 & 7.4 {\gdown{30.8\%}}
  & 1010 & 512.4 {\gdown{49.3\%}}
  & 121 & 85 {\gdown{29.8\%}} \\
SmolVLA
  & 450M & 354M {\gdown{21.3\%}}
  & 100M & 63M {\gdown{37\%}}
  & 24.75 & 8.83 {\gdown{64.3\%}}
  & 598.4 & 536.1 {\gdown{10.41\%}}
  & 201 & 137 {\gdown{31.84\%}} \\
\bottomrule
\end{tabular}
}
\vspace{-4mm}
\end{table}

\paragraph{RQ1. Compression Trade-Offs.}
We first investigate the limits of layer removal across distinct model families. 
As demonstrated in Figure~\ref{fig:result_abcde}, both $\pi_0$ 
on LIBERO (a) and GR00T-N1.5 on RoboCasa (b) exhibit \textit{flat performance profiles up 
to a $50\%$ pruning ratio}. This capability allows us to cut total FLOPs by $\times 1.39$ 
and $\times 1.42$ with $\pi_0$ and GR00T-N1.5, respectively, with negligible loss in policy success rate. These 
findings validate our core hypothesis: a substantial fraction of deep VLA layers 
are functionally redundant and can be statically removed prior to fine-tuning. 

We also present in the Table \ref{tab:efficiency} an overall efficiency reported across several factors such as model size, trainable parameters, training time, GFLOPs, and inference speed across three VLA backbones $\pi_{0}$, GR00T-N1.5, and SmolVLA~\cite{shukor2025smolvla}. It can be seen that \texttt{CLP} demonstrates generalizability across structurally diverse VLA backbones, uniformly compressing total \textit{(i) model size} by 21.3\%–25.9\% and \textit{(ii) cutting trainable parameters} by 25.8\%–37.0\% from sub-billion to multi-billion scales as well as (iii) saving GFLOPs up to 50\% while preserving equivalent success rate (Table~\ref{tab:libero}). 

%%%%%%%%%%%%%%%%%%%%%%%%%%%
% \vspace{-0.1in}
\paragraph{RQ2. Latent Behavior \& Ablation.}
To understand this performance recovery, we analyze the hidden-state geometry of the compressed model using PCA (Figure~\ref{fig:method}, top-right). Before fine-tuning, structural pruning severely contracts the latent space, collapsing representations into a narrow subspace relative to the diverse distribution of the base model. After target-task adaptation, however, the remaining layers reorganize their feature pathways, restoring a representation manifold closely aligned with the original network. This geometric “manifold restoration” explains how the compressed policy regains expressive capacity while maintaining baseline-level manipulation performance.

Subsequently, we explore alternative block-selection CKA, including (i) Mean-squared error (\textsc{MSE}); (ii) \textsc{Cosine} similarity; (iii) \textsc{random}, where layers are selected uniformly at random, and finally (iv) \textsc{keep-first}, in which the last $k$ layers are removed, retaining only the earliest layers of the network. As shown in Figure~\ref{fig:result_abcde}-d, \textsc{CKA} consistently delivers the most stable performance across all benchmarks, closely matching the unpruned baseline while maintaining the highest average success rate. In contrast, localized similarity metrics and heuristic baselines can produce more unstable degradation, particularly on long-horizon and spatial tasks. Hidden-state analysis in Figure\ref{fig:result_abcde}-f further reveals that these alternatives distort representations into isolated subspaces, whereas CKA preserves the global topology necessary for effective post-pruning adaptation. 

% \Ngan{MSE and CKA deliver similar results. What is the computation cost comparison between MSE and CKA? }

%%%%%%%%%%%%%%%%%%%%%%%%%%%
% \begin{figure}[!htb]
%     \centering
%     \includegraphics[width=\linewidth]{images/compare_mole.pdf}
%     \caption{Comparision with dynamic layers skipping method vs pruning}
%  \label{fig:method}
% \end{figure}

% \begin{wrapfigure}{r}{0.5\textwidth}
%     \centering
%     \includegraphics[width=\linewidth]{images/compare_mole.pdf}
%     \caption{Comparison with dynamic layer skipping method vs pruning}
%     \label{fig:method}
% % \end{wrapfigure}

\begin{wrapfigure}{r}{0.5\textwidth}
    \vspace{-15pt}          % pull figure up if there's space above
    \centering
    \includegraphics[width=\linewidth]{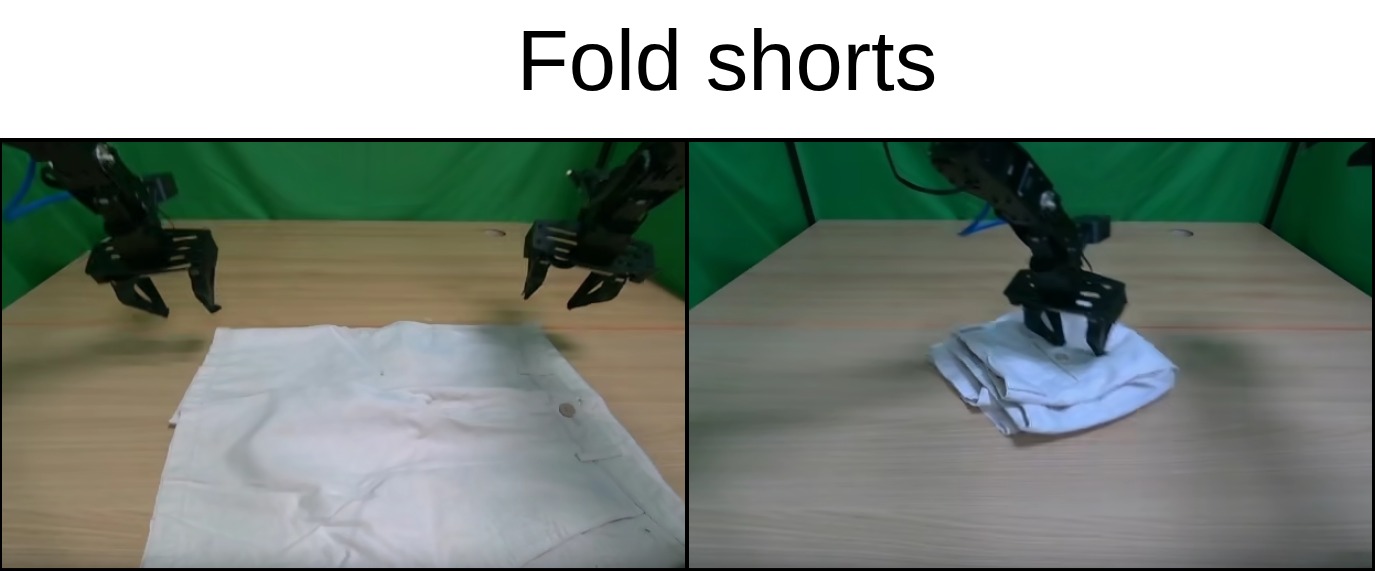}
    \vspace{-0.1in}
    \caption{Robot experiments with folding shorts.}
    \label{fig:real-world-exp}
    \vspace{-0.1in}          % shrink the dead space below the figure
\end{wrapfigure}
% \vspace{-0.1in}
\paragraph{RQ3. State-of-the-art  Compression Comparison.}
To evaluate the practical utility of \texttt{\texttt{CLP}}, we (i) compare with training-free pruning across multiple VLA settings. 
On LIBERO (Table~\ref{tab:libero}), \textsc{CLP} consistently achieves a superior efficiency--performance trade-off, delivering 1.39--1.47$\times$ speedups while maintaining near-baseline success rates across three modern VLA backbones. Furthermore, it is important to note that these token-pruning methods primarily accelerate inference and \textit{do not reduce the cost of downstream fine-tuning}, which remains a major bottleneck for VLA adaptation.
% On LIBERO (Table~\ref{tab:libero}), \texttt{\texttt{CLP}} consistently outperforms training-free pruning approaches without any pruning visual token as others. 

We further benchmark against (ii) trainable MoLe-VLA~\cite{zhang2026mole} in a challenging few-shot setting using only 10\% of the LIBERO training data (Fig.~\ref{fig:result_abcde}-c). Despite its simplicity, \texttt{\texttt{CLP}} achieves an 84.6\% average success rate, surpassing both the full $\pi_0$ baseline (77.7\%) and $\pi_0$-MoLe (79.7\%), while reducing training time by 1.38$\times$. Finally, on (iii) SimplerEnv with GR00TN1.5 (Tab.~\ref{tab:simpler_GR00T}), \texttt{\texttt{CLP}} improves the average success rate from 16.6\% to 20.0\% while shortening training time from 22.9 to 15.7 hours. These results show that structure-aware compression coupled with lightweight fine-tuning provides a more effective and scalable alternative to both inference-only pruning and parameter-expanding adaptive architectures, particularly in low-data regimes.

\vspace{-0.05in}
\begin{table}[h]
\centering
\vspace{-0.15in}
\caption{LIBERO benchmark comparison with training-free acceleration methods.}
\vspace{0.05in}
\resizebox{0.95\textwidth}{!}{
\begin{tabular}{lcccccc}
\toprule
\multirow{2}{*}{\textbf{Method}} &
\multicolumn{4}{c}{\textbf{Success Rate (\%)}} &
\textbf{Avg. SR} &
\textbf{Speedup} $\uparrow$ \\
\cmidrule(lr){2-5}
& Spatial & Object & Goal & Long &
\textbf{(\%)} & \\
\midrule

OpenVLA-OFT~\cite{kim2025fine}
& 97.6\% & 96.5\% & 97.9\% & 94.5\%
& 96.6\% & $1.00\times$ \\

FastV~\cite{chen2024image}
& 94.6\% & 95.8\% & 94.0\% & 88.8\%
& 93.3\% & $1.44\times$ \\

DivPrune~\cite{alvar2025divprune}
& 92.4\% & 91.2\% & 89.0\% & 84.8\%
& 89.4\% & $1.46\times$ \\

EfficientVLA~\cite{yang2026efficientvla}
& 96.5\% & 91.1\% & 96.0\% & 72.1\%
& 88.9\% & $1.52\times$ \\

ADP~\cite{pei2025action}
& 97.6\% & 98.4\% & 97.4\% & 84.2\%
& 94.4\% & $1.35\times$ \\

\midrule

$\pi_0$
& 94.6\% & 98.2\% & 95.4\% & 90.0\%
& 94.6\% & $1.00\times$ \\

$\pi_0$-SpecPrune-VLA~\cite{wang2025spec}
& 96.6\% & 98.0\% & 95.2\% & 84.2\%
& 93.5\% & $1.31\times$ \\

\rowcolor{rowcolor}
$\pi_0$-CLP \textbf{(Ours)}
& 95.0\% & 99.2\% & 95.0\% & 86.4\%
& 93.9\% & $1.39\times$ \\

\midrule

GR00T-N1.5
& 90.8\% & 98.4\% & 95.4\% & 91.0\%
& 93.9\% & $1.00\times$ \\

\rowcolor{rowcolor}
GR00T-N1.5-CLP \textbf{(Ours)}
& 89.4\% & 98.8\% & 95.8\% & 88.6\%
& 93.0\% & $1.42\times$ \\

\midrule

SmolVLA
& 71.8\% & 92.2\% & 87.4\% & 57.2\%
& 77.15\% & $1.00\times$ \\

\rowcolor{rowcolor}
SmolVLA-CLP \textbf{(Ours)}
& 75.6\% & 93.0\% & 81.6\% & 56.2\%
& 76.75\% & $1.47\times$ \\

\bottomrule
\end{tabular}}
\vspace{0.05in}
\label{tab:libero}
\end{table}

\newcolumntype{C}[1]{>{\centering\arraybackslash}p{#1}}
\newcommand{\gain}[2]{%
  $\hspace{0.22in} \textbf{#1}_{\textcolor{green!60!black}{\uparrow#2}}$%
}
\newcommand{\second}[2]{%
  $\hspace{0.22in} \underline{#1}_{\textcolor{green!60!black}{\uparrow#2}}$%
}
\begin{table}[!htb]
\vspace{-0.05in}
\small
\vspace{-0.1in}
\renewcommand{\arraystretch}{0.9}\setlength{\tabcolsep}{3pt}
\setlength{\tabcolsep}{1pt}
\centering
\caption{\textbf{Evaluation of GR00T N1.5 on SimplerEnv}}
\vspace{0.05in}
\label{tab:simpler_GR00T}
\resizebox{0.92\textwidth}{!}{
\begin{tabular}{c|c|ccccccc|c}
\toprule
\multirow{3}{*}{\textbf{Model}} &
\multirow{3}{*}{\textbf{\makecell{Training\\Time\\(hours)}}} &
\multicolumn{7}{c|}{\textbf{WidowX}} & \textbf{Task(All)} \\
\cmidrule(lr){3-9} \cmidrule(lr){10-10}
& & \makecell{Carrot\\Plate}
& \makecell{Eggplant\\Basket}
& \makecell{Spoon\\Towel}
& \makecell{Stack\\Cube}
& \makecell{Eggplant\\Sink}
& \makecell{Close\\Drawer}
& \makecell{Open\\Drawer}
& \makecell{\textbf{Avg.}} \\
\midrule
GR00TN1.5        & 22.9 & 26 & \textbf{34} & 18  & \textbf{8} & 8 & 12 & 10 & 16.57 \\
\rowcolor{rowcolor}
GR00TN1.5-CLP  & 15.7 & \textbf{34} & 14 & \textbf{38} & 4 & \textbf{16} & \textbf{24} & \textbf{10} & \textbf{20} \\
\bottomrule
\end{tabular}}
\vspace{-4mm}
\end{table}
%%%%%%%%%%%%%%%%%%%%%%%%%%%
\vspace{0.15in}
\paragraph{RQ4. Real-World Deployment.}
We validate \texttt{\texttt{CLP}} through 10 real-world manipulation tasks spanning four robotic embodiments (UR10, UR5, single-arm ALOHA, and bimanual ALOHA) (Figures.~\ref{fig:method} and ~\ref{fig:real-world-exp}). Despite reducing model depth, \texttt{\texttt{CLP}} maintains or improves overall task performance (Tab.~\ref{tab:GR00T15_tasks}), increasing the average success rate from 73.5\% to 75.9\% on GR00TN1.5 while outperforming the full model on several challenging tasks, including napkin serving (+20\%), screwdriver placement (+15\%), banana-to-pot transfer (+10\%), and bimanual cloth folding (+5\%). Importantly, these gains are accompanied by substantial reductions in fine-tuning time, achieving up to 1.94$\times$ faster training in physical deployments (Fig.~\ref{fig:result_abcde}-e and Appendix.)

% We argue that this enhancement can be reasoned via \textit{implicit regularization} where permanently removing structural redundancies acts as a robust regularizer that mitigates overfitting to training noise. Furthermore, fine-tuned pruned models can force the remaining layers to rebuild expressivity and tightly reoccupy the multi-modal latent space distribution of the base network. In summary,
% the consistency of these results across diverse manipulation skills, object types, and robot embodiments confirms that \texttt{\texttt{CLP}} preserves real-world control capability while delivering tangible efficiency benefits, validating its practicality for large-scale robotic adaptation beyond simulation.

We hypothesize that these gains arise from an implicit regularization effect: removing redundant layers reduces model capacity and discourages overfitting to task-specific noise. During fine-tuning, the remaining layers reorganize to recover the expressive latent structure of the original network, often yielding better adaptation under limited data. The consistency of these improvements across diverse tasks, objects, and robot embodiments demonstrates that \texttt{CLP} enhances efficiency without compromising real-world control capability.
\vspace{-0.1in}
\begin{table}[!htb]
\vspace{-0.05in}
\small
\renewcommand{\arraystretch}{0.9}
\setlength{\tabcolsep}{1pt}
\centering
\caption{\textbf{Performance on Real-world manipulation tasks (GR00TN1.5).} Task success rate (\%).}
\label{tab:GR00T15_tasks}
\resizebox{1.0\textwidth}{!}{%
\begin{tabular}{c|ccc|cc|ccc|cc|c}
\toprule
\multirow{3}{*}{\textbf{Model}}
  & \multicolumn{8}{c|}{\textbf{Single Arm}}
  & \multicolumn{2}{c|}{\textbf{Bimanual}}
  & \multirow{3}{*}{\textbf{Avg.}} \\
\cmidrule(lr){2-9} \cmidrule(lr){10-11}
  & \multicolumn{3}{c|}{\textbf{UR10}}
  & \multicolumn{2}{c|}{\textbf{UR5}}
  & \multicolumn{3}{c|}{\textbf{ALOHA Single Arm}}
  & \multicolumn{2}{c|}{\textbf{ALOHA Bimanual}} \\
\cmidrule(lr){2-4} \cmidrule(lr){5-6} \cmidrule(lr){7-9} \cmidrule(lr){10-11}
  & \makecell{Groceries\\ToBasket}
  & \makecell{Open\\Kettle}
  & \makecell{Close\\Kettle}
  & \makecell{Serve\\Napkin}
  & \makecell{Screwdriver\\ToBasket}
  & \makecell{Banana\\ToPot}
  & \makecell{Cube\\ToDrawer}
  & \makecell{Block\\Stacking}
  & \makecell{Fold\\Shorts}
  & \makecell{Fly\\Towel}
  & \\
\midrule
GR00TN1.5  & \textbf{90} & \textbf{100} & \textbf{100} & 45 & 15 & 65 & \textbf{75} & \textbf{80} & 90 & \textbf{75} & 73.5 \\
\rowcolor{rowcolor}
Gr00N1.5-CLP & 89 & 95 & \textbf{100} & \textbf{65} & \textbf{30} & \textbf{75} & 60 & 75 & \textbf{95} & 70 & \textbf{75.9} \\
\bottomrule
\end{tabular}%
}
\vspace{-2mm}
\end{table}
\section{Conclusion}
% \vspace{-0.1in}
\label{sec:conclusion}
% We have shown that a simple yet effective CKA-guided Layer Pruning strategy can substantially simplify state-of-the-art VLA models, including $\pi_0$ and GR00T-N1.5, without relying on complex architectural modifications or adaptive routing mechanisms. Through extensive simulation and real-world evaluations, we demonstrate that these VLA ones exhibit significant structural redundancy, enabling the removal of 30--50\% of transformer layers while maintaining competitive performance. Beyond matching or surpassing existing acceleration techniques, our \textsc{\texttt{CLP}} consistently improves training efficiency and exhibits strong advantages in low-data and real-world robot setups. We hope this work encourages the community to rethink the necessity of ever-growing model architectures and inspires a new generation of efficient, scalable, and environmentally sustainable VLA systems.
\vspace{-0.1in}
We demonstrate that a simple yet effective \textit{CKA-guided Layer Pruning} strategy can substantially simplify state-of-the-art VLA models, including $\pi_0$ and GR00T-N1.5, without architectural modifications or adaptive routing mechanisms. Across extensive simulation and real-world evaluations, we show that modern VLA backbones contain significant structural redundancy, allowing 30--50\% of transformer layers to be removed while maintaining competitive performance. Beyond achieving a favorable efficiency--performance trade-off, \texttt{CLP} consistently accelerates adaptation and exhibits strong gains in low-data and real-world robotic settings. We hope these findings encourage the community to rethink the necessity of ever-growing VLA architectures and inspire more efficient, scalable, and sustainable foundation policies.

% ============================================================
\vspace{-0.1in}
% \paragraph{Limitation.}
\section{Limitation}
\label{sec:limitation}
\vspace{-0.1in}
A key limitation of \textsc{CLP} is its use of a global pruning criterion that does not explicitly account for modality-specific token dynamics in manipulation tasks. Our analysis suggests that action and state tokens exhibit distinct representations, motivating future token- or modality-aware pruning strategies. In addition, the current work investigates \textsc{CLP} exclusively in the context of post-pretraining fine-tuning, leaving its application to the pretraining stage an open direction for future work. Rather than relying on heuristic layer selection as in existing approaches, \textsc{CLP} offers a principled mechanism for identifying which layers of a pretrained VLM are most beneficial to retain or adapt when transitioning to a VLA, suggesting its potential as a layer-selection prior during pretraining-stage adaptation.

% \clearpage
% \newpage

% \section{Experimental Evaluation}

% \label{sec:06_CKA-guide_pruning}
% \input{sections/06_CKA-guide_pruning}

% \section{Discussion}
% \label{sec:07_discussion}
% \input{sections/07_discussion}

% \section{Conclusion}
% \label{sec:08_conclusion}
% \input{sections/08_conclusion}

% \textcolor{red}{Todo 28/05
% \begin{itemize}
%     \item Update Introduction \textcolor{green}{done}
%     \item Update Related Works \textcolor{green}{done}
%     \item Re-write the method section  \textcolor{orange}{(in progress)}
%     \item Add LIBERO training-free table $\rightarrow$ \textcolor{green}{done}
%     \item Evaluate skipping (MoLe VLA) on pi0 LIBERO 10\% and ROBOCASA \textcolor{green}{(done)}
%     \item Draw method figure 1 \textcolor{green}{(done)}
%     \item Draw plot figure (SR vs FLOPs, SR vs training time) \textcolor{green}{(done)}
%     \item Remove cosine similarity figure 2 \textcolor{green}{(done)}
% \end{itemize}}

% % ============================================================
% \clearpage
% \acknowledgments{
% % Uncomment for camera-ready:
% % We thank ...
% }

% % ============================================================
% No \bibliographystyle needed — corl_2026.sty handles it
\bibliography{references}

% % ============================================================

\appendix
\newpage

% Add only appendix entries to TOC
\addtocontents{toc}{\protect\setcounter{tocdepth}{2}}

\section*{Supplement to "Finetuning Vision-Language-Action Models Requires
Fewer Layers Than You Think"}
\addcontentsline{toc}{section}{Appendix}

% Mini table of contents for appendix only
\startcontents[appendix]
\printcontents[appendix]{l}{1}{\setcounter{tocdepth}{2}}
% \newpage
\section{Appendix}
% \section{Appendix}

\subsection{Implementation details}

To systematically determine the redundancy across transformer layers, we implemented a data-driven pruning pipeline across both simulated and real-world environments. We sampled data batches from diverse datasets and recorded the intermediate hidden states across all network layers during the forward pass. Utilizing these representations, we computed Centered Kernel Alignment (CKA) matrices to quantify the representational similarity between different transformer layers. 

\begin{table}[h]
\centering
\caption{Summary of pruning configurations and layer distributions across the evaluated VLA models.}
\resizebox{\columnwidth}{!}{%
\begin{tabular}{llccl}
\hline
\textbf{Model} & \textbf{Module} & \textbf{\begin{tabular}[c]{@{}c@{}}Number of transformer\\layers in original model\end{tabular}} & \textbf{\begin{tabular}[c]{@{}c@{}}Number of transformer\\layers pruned\end{tabular}} & \textbf{Pruned layer index} \\
\hline
$\pi_0$ & VLM and Action expert & 18 & 12 & 1, 2, 4, 6, 8, 9 \\
\hline
\multirow{3}{*}{GR00T-N1.5} & VLM & 12 & 5 & 3, 4, 5, 6, 7, 8, 9 \\
 & VL-self-attention & 4 & 3 & 2 \\
 & DiT Action head & 16 & 8 & 1, 2, 4, 5, 6, 7, 10, 11 \\
\hline
SmolVLA & VLM and Action expert & 16 & 10 & 1, 2, 5, 6, 14, 15 \\
\hline
\end{tabular}%
}
\label{tab:pruning_config}
\end{table}

Based on the resulting CKA similarity profiles, we defined a threshold, $\tau$, to segment the network into distinct CKA blocks, as illustrated in Figure~\ref{fig:method}. For each identified block $K$, we retained only the first layer-hypothesizing that the initial layer is critical for processing and establishing the block's input representations - and pruned all subsequent layers within that same block. This structured pruning methodology was uniformly applied to three popular VLA models: $\pi_0$, GR00T-N1.5, and SmolVLA. The specific architectural configurations and post-pruning layer distributions for each model are summarized in Table~\ref{tab:pruning_config}.

% \begin{algorithm}[t]
\begin{algorithm}[H]
\caption{CKA-Guided Layer Pruning}
\label{alg:cka_pruning}
\begin{algorithmic}[1]
\Require Pre-trained policy $\pi_\theta$, target module $\mathcal{M}$, calibration set $\mathcal{D}_{\mathrm{cal}}$, budget $k_{\mathcal{M}}$, threshold $\tau$
\State Extract calibrated hidden representations $\{\bar{H}^{\mathcal{M}}_1,\ldots,\bar{H}^{\mathcal{M}}_{L_{\mathcal{M}}}\}$ from $\mathcal{M}$ over $\mathcal{D}_{\mathrm{cal}}$
\For{$\ell=2$ to $L_{\mathcal{M}}$}
    \State $s^{\mathcal{M}}_\ell \leftarrow \operatorname{CKA}(\bar{H}^{\mathcal{M}}_{\ell-1},\bar{H}^{\mathcal{M}}_{\ell})$
\EndFor
\State Group consecutive layers into blocks $\mathcal{B}_{\mathcal{M}}$ where $s^{\mathcal{M}}_\ell\geq\tau$
\State Identify candidate pool $\mathcal{P}_{\mathcal{M}}\leftarrow \bigcup_{B\in\mathcal{B}_{\mathcal{M}}}(B\setminus\{r(B)\})$, where $r(B)$ is the initial layer of block $B$
\State Select removal set $\mathcal{R}_{\mathcal{M}}\leftarrow \operatorname{TopK}_{\ell\in\mathcal{P}_{\mathcal{M}}}(s^{\mathcal{M}}_\ell,k_{\mathcal{M}})$
\State $\pi_\theta^{\mathrm{pruned}}\leftarrow \operatorname{RemoveLayers}(\pi_\theta,\mathcal{R}_{\mathcal{M}})$
\State \Return $\pi_\theta^{\mathrm{pruned}}$
\end{algorithmic}
\end{algorithm}

\subsection{Additional Comparisons with Adaptive Training Compression Methods}

In this section, we further compare CLP against training-adaptive baselines to demonstrate that CLP not only accelerates inference but also significantly reduces training cost. To ensure a fair comparison, we integrate dynamic layer skipping strategy from MoLe-VLA~\cite{zhang2026mole} on $\pi_0$ and train all models for the same number of steps. Unlike MoLe, which introduces additional trainable modules for dynamic layer selection and thus incurs even longer training times, CLP operates within a fixed pruned architecture requiring no extra parameters. Experiments on 10\% of the LIBERO dataset and a 30-demonstration subset of ROBOCASA show that CLP achieves approximately 28\% and 23\% reduction in training time compared to the base model, respectively, while simultaneously improving average performance by 6.9\% on LIBERO and 2.4\% on ROBOCASA. These results suggest that pruning redundant layers not only streamlines inference but also reduces the optimization burden during fine-tuning, offering a more efficient training regime without sacrificing performance (Tables~\ref{tab:libero_scale_10} and~\ref{tab:pi0_robocasa}).

% \input{tables/pi0_robocasa}
% \begin{table}[h]
% \centering
% \caption{Libero Benchmark Results training on 10\% data (\% Success Rate)}
% \begin{tabular}{lccccc}
% \toprule
% \textbf{Model} & \textbf{Long} & \textbf{Goal} & \textbf{Object} & \textbf{Spatial} & \textbf{Average} \\
% \midrule
% $\pi_0$   & 58.8 & 87.8 & 82.6 & 81.6 & \textbf{77.7} \\
% $\pi_0$ - MoLe    & 60.2 & 88.2 & 86.0 & 84.4 & \textbf{79.7} \\
% \rowcolor{rowcolor} $\pi_0$-CLP (Ours)       & 66.2 & 90.6 & 89.0 & 92.6 & \textbf{84.6} \\
% \bottomrule
% \end{tabular}
% \end{table}

\begin{table}[h]
\centering
\caption{Libero Benchmark Results training on 10\% data (\% Success Rate)}
\begin{tabular}{lcccccc}
\toprule
\textbf{Model} & \textbf{Long} & \textbf{Goal} & \textbf{Object} & \textbf{Spatial} & \textbf{Average} & \textbf{Training} \\
& & & & & & \textbf{Hours} \\
\midrule
$\pi_0$   & 58.8 & 87.8 & 82.6 & 81.6 & \textbf{77.7} & 15.5\\
$\pi_0$ - MoLe    & 60.2 & 88.2 & 86.0 & 84.4 & \textbf{79.7} & 15.6 \\
\rowcolor{rowcolor} $\pi_0$-CLP (Ours)       & 66.2 & 90.6 & 89.0 & 92.6 & \textbf{84.6} & 11.2\\
\bottomrule
\label{tab:libero_scale_10}
\end{tabular}
\end{table}

\subsection{Real-world experiments details}

In this work, we conduct experiments on 10 real-world tasks across 4 different robot embodiments. The collected tasks illustrated on Figure~\ref{fig:all_realworld} cover a diverse range of manipulation skills, from single-arm to bimanual manipulation, and from rigid to deformable object handling, spanning data scales from 100 to 2800 demonstrations. Details are summarized in Table~\ref{tab:real_robot_details}.

\begin{table}[h]
\centering
\caption{Real-world task details.}
\vspace{0.05in}
\label{tab:real_robot_details}
\resizebox{\textwidth}{!}{%
\begin{tabular}{l c l c p{3.0in}}
\toprule
Task & No. Episodes & Embodiment & No. Views & Task Description \\
\midrule
Groceries To Basket & 2800 & UR10e & 2 &
    Pick up the tomato can/ coffee bag/ garlic powder/ ketchup/ mayonnaise/mustard/ olive oil/spam can and place them in the grey bin. \\
Open Kettle & 300 & UR10e & 2 & Open the lid of the electric kettle. \\
Close Kettle & 300 & UR10e & 2 & Close the lid of the electric kettle. \\
Serve Napkin & 100 & UR5 & 2 & Pull napkin out from the box. \\
Screwdriver To Basket & 100 & UR5 & 2 & Pick up screwdriver and place in box. \\
Banana in Pot & 150 & Single-arm ALOHA & 2 & Use the right gripper to pick up the banana and place it into the pot.Then pick up the lid and place it on top of the pot to close it. \\
Cube To Drawer & 150 & Single-arm ALOHA & 2 & Use the right gripper to open the top drawer, pick up the red square block, place it inside the top drawer, and then close the drawer. \\
Block Stacking & 151 & Single-arm ALOHA & 2 & Pick up the yellow triangular block and place it on top of the orange square block. \\
Fold Shorts & 202 & Bimanual ALOHA & 3 & Use both grippers to fold the shorts by bringing one leg over the other,    then fold the bottom upward to form a compact shape. \\
Fly Towel & 101 & Bimanual ALOHA & 3 & Use both grippers to grasp two corners of the towel, lift it, and fling    it onto the table so that it spreads out flat. \\
\bottomrule
\end{tabular}
}
\end{table}

\begin{figure}[!h]
    \centering
    \includegraphics[width=\linewidth]{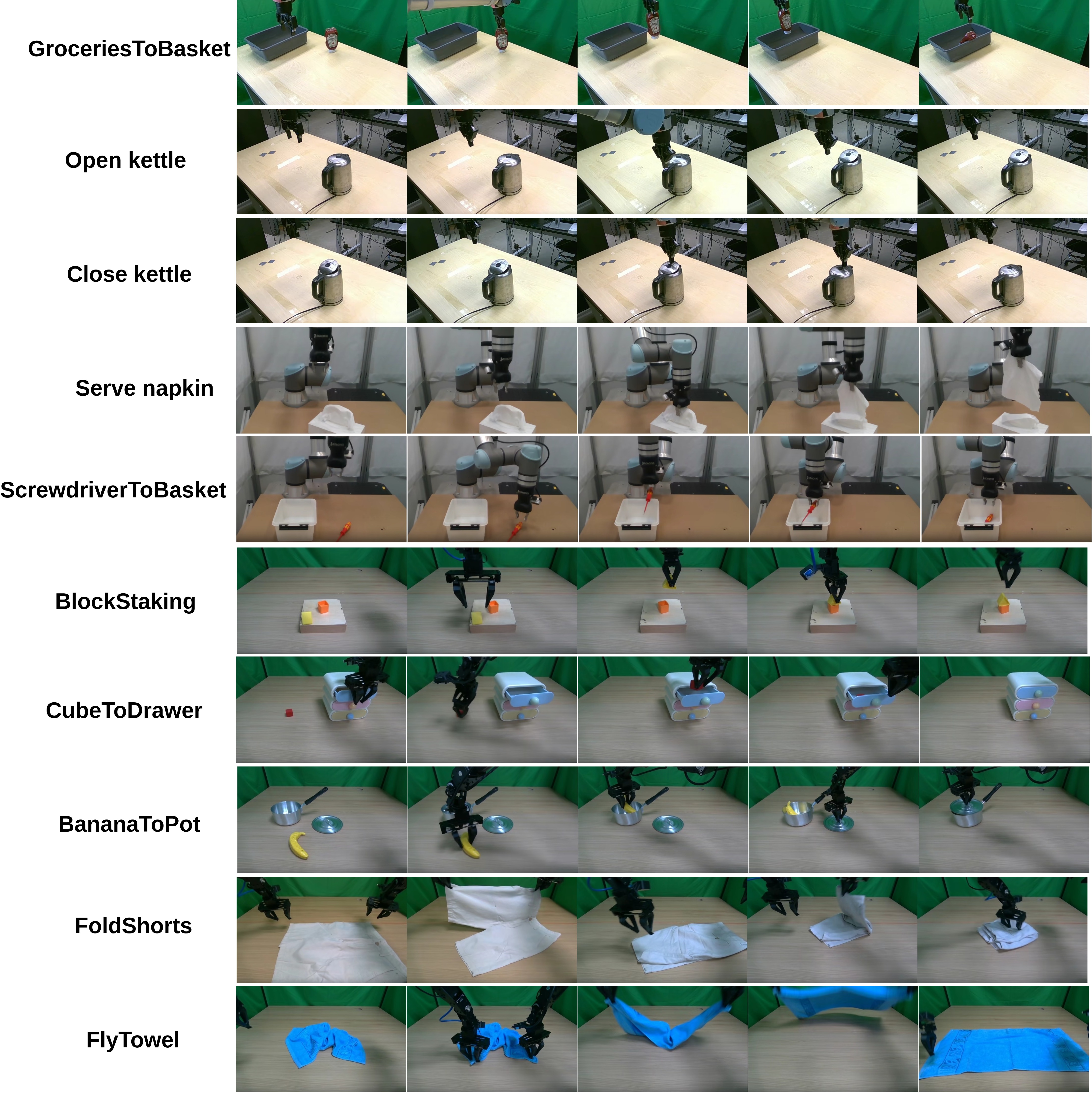}
    \caption{Real-world data samples}
 \label{fig:all_realworld}
\end{figure}

We fine-tuned both GR00T-N1.5 and GR00T-N1.5-CLP on a single NVIDIA H100 GPU with a batch size of 32. All hyperparameters were kept at their default values except \texttt{MAX\_STEP}. Training steps and training time are reported in Table~\ref{tab:training_comparison}, and evaluation results are presented in Table~\ref{tab:GR00T15_tasks}.

% \begin{table}[h]
% \centering
% \caption{Training step and training time  on real-world datasets}
% \label{tab:training_comparison}
% \begin{tabular}{l c c c c}
% \toprule
% \multirow{2}{*}{\textbf{Task}} & \multicolumn{2}{c}{\textbf{GR00T-N1.5}} & \multicolumn{2}{c}{\textbf{GR00T-N1.5-CLP}} \\
% \cmidrule(r){2-3} \cmidrule(l){4-5}
% & \textbf{Training step} & \textbf{Training time (hours)} & \textbf{Training step} & \textbf{Training time (hours)} \\
% \midrule
% GroceriesToBasket   & 100k & 11.8 & 100k & 8   \\
% OpenKettle          & 24k  & 2.8  & 18k  & 1.4 \\
% CloseKettle         & 26k  & 3    & 19k  & 1.5 \\
% ServeNapkin         & 10k  & 1.1  & 10k  & 0.7 \\
% ScrewdriverToBasket & 13k  & 1.1  & 13k  & 1.5 \\
% BananaToPot         & 30k  & 5.1  & 25k  & 2.9 \\
% CubeToDrawer        & 33k  & 5.6  & 28k  & 3.2 \\
% BlockStacking       & 24k  & 2.8  & 18k  & 1.4 \\
% FoldShort           & 45k  & 6.5  & 45k  & 4.4 \\
% FlyTowel            & 13k  & 3.2  & 13k  & 2.1 \\
% \bottomrule
% \end{tabular}
% \end{table}

\begin{table}[h]
\centering
\caption{Training step and training time on real-world datasets}
\label{tab:training_comparison}
\resizebox{\textwidth}{!}{%
\begin{tabular}{l c c c c}
\toprule
\multirow{2}{*}{\textbf{Task}} & \multicolumn{2}{c}{\textbf{GR00T-N1.5}} & \multicolumn{2}{c}{\textbf{GR00T-N1.5-CLP}} \\
\cmidrule(r){2-3} \cmidrule(l){4-5}
& \textbf{Training step} & \textbf{Training time (hours)} & \textbf{Training step} & \textbf{Training time (hours)} \\
\midrule
GroceriesToBasket   & 100k & 11.8 & 100k & 8   \\
OpenKettle          & 24k  & 2.8  & 18k  & 1.4 \\
CloseKettle         & 26k  & 3    & 19k  & 1.5 \\
ServeNapkin         & 10k  & 1.1  & 10k  & 0.7 \\
ScrewdriverToBasket & 13k  & 1.5  & 13k  & 1.1 \\
BananaToPot         & 30k  & 5.1  & 25k  & 2.9 \\
CubeToDrawer        & 33k  & 5.6  & 28k  & 3.2 \\
BlockStacking       & 24k  & 2.8  & 18k  & 1.4 \\
FoldShorts           & 45k  & 6.5  & 45k  & 4.4 \\
FlyTowel            & 13k  & 3.2  & 13k  & 2.1 \\
\bottomrule
\end{tabular}%
}
\end{table}

\begin{figure}[!h]
    \centering
    \includegraphics[width=\linewidth]{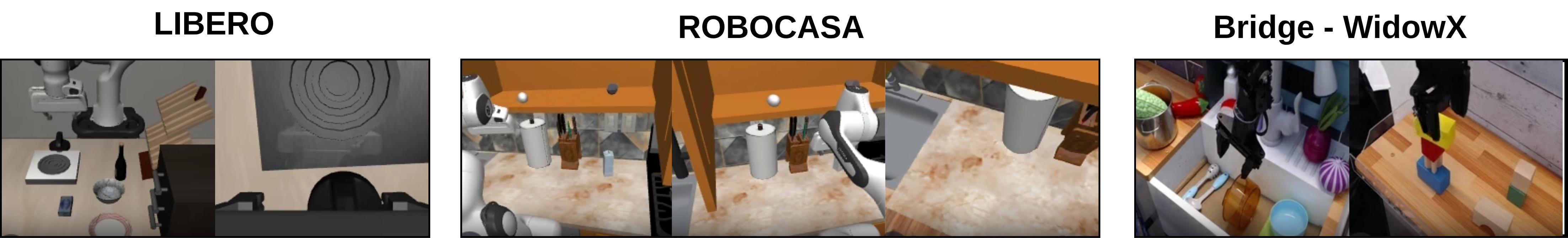}
    \caption{Simulation data samples}
 \label{fig:all_realworld}
\end{figure}

\subsection{Simulation experiments details}

\textbf{LIBERO:} We use LIBERO as a benchmark to evaluate our method against other baselines. A 10\% subset is randomly sampled from the full LIBERO dataset. We fine-tune all models - $\pi_0$, GR00T-N1.5, and SmolVLA - on this subset for 100k steps with a global batch size of 64. We evaluate each task over 50 episodes with an execution length of 10 steps.Results are summarized in Table~\ref{tab:libero}.

\textbf{ROBOCASA:} We used small subsets 30 demos and 100 demos of ROBOCASA and sampled demonstrations from 5 tasks covering diverse skills: \textit{PnPCabToCounter}, \textit{PnPCounterToCab}, \textit{SetUpCoffeeMug}, \textit{TurnOffStove}, and \textit{TurnOnMicrowave}, drawn from the original 24-task dataset. We train both baseline and pruned models on this subset for 100k steps. The global batch size for $\pi_0$ is 48, fine-tuned on 4 H100 GPUs, while GR00T-N1.5 uses a global batch size of 32 on a single GPU. We evaluate each task over 50 episodes with an execution length of 10 steps. The results shown in Figure~\ref{fig:result_abcde}-b and Table~\ref{tab:pi0_robocasa}

\begin{table}[H]
\centering
\caption{Robocasa 30 demos results(\% Success Rate)}
\resizebox{\textwidth}{!}{%
\begin{tabular}{lccccccc}
\toprule
\textbf{Model} & 
\textbf{PnP Cab} & 
\textbf{PnP Counter} & 
\textbf{Coffee Setup} & 
\textbf{Turn Off} & 
\textbf{Turn On} & 
\textbf{Average} & 
\textbf{Training} \\
& 
\textbf{to Counter} & 
\textbf{to Cab} & 
\textbf{Mug} & 
\textbf{Stove} & 
\textbf{Microwave} & 
& 
\textbf{time (hours)} \\
\midrule
$\pi_0$ base  & 14 & 16 & 2 & 2 & 44 & 15.6 & 17.5 \\
$\pi_0$- MoLe        & 14 & 18 & 2 & 4 & 50 & 17.6 & 17.7 \\
\rowcolor{rowcolor} $\pi_0$-CLP (Ours)        & 16 & 16 & 4 & 4 & 50 & 18 & 13.5 \\
\bottomrule
\end{tabular}}
\label{tab:pi0_robocasa}
\end{table}

\textbf{SimplerEnv:} We fine-tune both the base and pruned GR00T-N1.5 models on the Bridge dataset on a single H100 GPU with  batch size of 32 for 200k steps, then evaluate on SimplerEnv WidowX tasks. We evaluate each task over 50 episodes with an execution length of 8 steps. Results are summarized in Table~\ref{tab:simpler_GR00T}.

\label{sec:appendix}

\end{document}